\documentclass[runningheads]{llncs}

\usepackage{eccv}

\usepackage{eccvabbrv}

\usepackage{graphicx}
\usepackage{booktabs}

\usepackage[accsupp]{axessibility}  % Improves PDF readability for those with disabilities.

\usepackage{comment}

\usepackage{hyperref}

\usepackage{orcidlink}

\begin{document}

\title{MVTrack: Ultrafast Appearance-Free Moving Object Tracking from Compressed Bitstreams} 

\titlerunning{MVTrack}

\author{Iñaki Erregue\inst{1,2,4}\orcidlink{0009-0004-1943-5554} \and
Kamal Nasrollahi\inst{3,4}\orcidlink{0000-0002-1953-0429} \and
Sergio Escalera\inst{1,2,3}\orcidlink{0000-0003-0617-8873}}

\authorrunning{I.~Erregue \etal}

%\institute{Princeton University, Princeton NJ 08544, USA \and
%Springer Heidelberg, Tiergartenstr.~17, 69121 Heidelberg, Germany

\institute{Universitat de Barcelona \and
Computer Vision Center \and Aalborg Universitet
\and Milestone Systems}

\maketitle

\begin{abstract}
  Deploying modern video trackers at scale is bottlenecked by the computational cost of RGB-based object detectors. To this end, we present MVTrack, an ultrafast tracker for moving objects that operates directly on H.264 bitstreams. MVTrack combines MVDet, a lightweight detector for motion vector fields, with MVLink, a minimalist kinematic association module. On VIRAT, MVTrack outperforms YOLO26n while using 60$\times$ fewer parameters, requiring 40$\times$ fewer FLOPs, and reducing CPU latency by 8.6$\times$. These results demonstrate that compressed video data alone can enable accurate and scalable surveillance tracking, thereby bypassing the need for pixel reconstruction.
  \keywords{Compressed-domain video analysis \and Motion vectors \and Multi-object tracking}
\end{abstract}

\section{Introduction}
Video surveillance has reached a planetary scale, with hundreds of millions of cameras monitoring urban, transportation, and industrial environments. Yet real-time analytics remains constrained by the cost of processing visual data, especially when many high-resolution feeds must run simultaneously on resource-limited edge hardware. While modern tracking-by-detection frameworks (TbD) achieve impressive accuracy, they typically rely on GPU-accelerated object detectors that operate on decoded RGB frames. As a result, CPU-only deployments must aggressively subsample frames, reduce resolution, or offload computation.

A key observation motivates this work: for many surveillance applications, appearance is not the primary signal of interest; motion is. Tasks such as traffic monitoring, intrusion detection, occupancy estimation, and trajectory-based anomaly detection ultimately depend on where objects move and how they move, rather than on detailed visual appearance. Yet most tracking systems implicitly assume that visual content must first be reconstructed before meaningful tracking can occur. Consequently, the dominant paradigm remains RGB-centric, even when the objective is simply to localize and associate moving entities over time.

Compressed video representations offer an alternative perspective. Modern video codecs such as H.264 encode predictive frames using motion vectors (MVs). Prior work has successfully exploited these compact displacement fields for object detection~\cite{9010861, 10350828, jimaging9070132, article_moura, POPPE2009428, huang2025mvp}, tracking~\cite{article, Alvar2018MVYOLOMV, 9248145, hwang2022cova, Nguyen2020TowardSV, 8575254, 7478003, elkhoury}, segmentation~\cite{9251974, 9879197, 8982035, 1294953,Hu_2023_CVPR, SOLANACIPRES200999}, action recognition~\cite{11259157, wu2018compressed, 7780666}, multi-modal video understanding~\cite{11095001, sarkar2026cope}, and numerous other vision tasks~\cite{8397016, gronquist2023efficient, zhou2023mvflow, 10.1145/3394171.3413504, 9509352, 7742914}. However, nearly all compressed-domain approaches retain a periodic RGB pathway. MVs accelerate inference on intermediate frames, while heavyweight RGB networks are invoked regularly to recover semantic information and correct accumulated errors. In other words, existing systems adopt a \textit{slow-fast} paradigm in which compressed-domain motion serves as a computational shortcut, but RGB remains the ultimate source of truth. We argue that, for moving-object tracking in surveillance settings, the \textit{slow} stream is unnecessary. Moving objects generate large, coherent MV patterns over time, while static scene content has near-zero displacement. Thus, MVs contain the signal needed to track moving entities. Rather than treating MVs as an auxiliary cue, we use them as the primary representation and eliminate RGB entirely. We call this the \textit{always-fast} paradigm.

Beyond computational efficiency, an always-fast approach offers distinct advantages for practical edge deployment. Because RGB frames are never reconstructed, the system provides a more privacy-aware representation by reducing direct exposure to biometric or appearance-based features. Furthermore, its minimal footprint enables real-time trajectory generation to summarize scene activity, serving as an efficient filter for event-based recording or anomaly detection. MVTrack can thus act as an always-on, low-power front end that triggers heavy, pixel-level perception models only when meaningful activity is detected. By bypassing both RGB decoding and appearance-based inference, this framework is well suited to resource-constrained, CPU-only edge devices and large-scale, multi-stream surveillance deployments.

To this end, we present \textbf{MVTrack}, an ultrafast moving-object tracker that operates entirely on H.264 bitstreams. We focus on pedestrians and vehicles, as these categories constitute most objects of interest in practical CCTV surveillance applications. The proposed TbD framework couples \textbf{MVDet}, a lightweight CenterNet-inspired~\cite{zhou2019objects} detector adapted to MV representations, with \textbf{MVLink}, a ByteTrack-based~\cite{zhang2022bytetrack} association module that uses kinematics to reduce identity fragmentation during stationary periods. Together, these components enable real-time tracking across multiple concurrent video streams on a single consumer-grade CPU while maintaining competitive accuracy. Tracking experiments on VIRAT~\cite{virat} demonstrate that MVTrack outperforms YOLO26n~\cite{yolo26_ultralytics} while using 60$\times$ fewer parameters, requiring 40$\times$ fewer FLOPs, and reducing CPU latency by a factor of 8.6$\times$. More broadly, our results suggest that useful surveillance tracking can be achieved without ever reconstructing a pixel, opening a path toward scalable, privacy-preserving, and resource-efficient video analytics.

\section{Related Work}
\subsection{Video Coding Preliminaries}
\begin{figure}[tb]
  \centering
  \includegraphics[width=\linewidth]{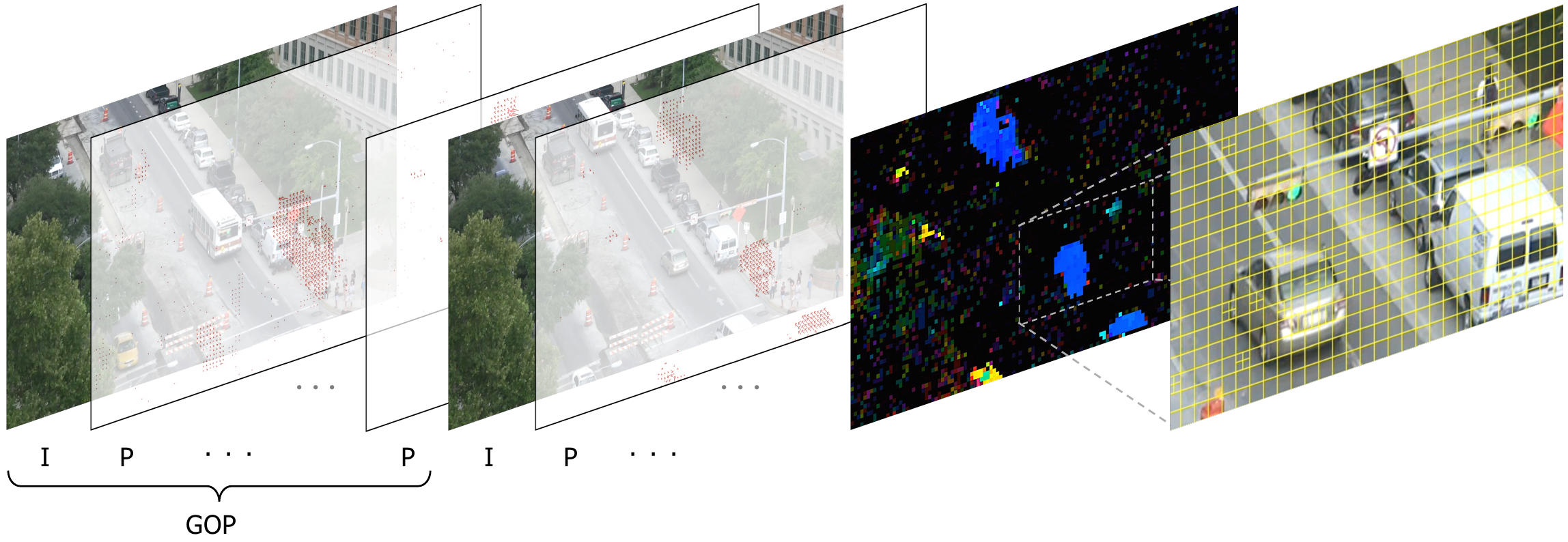}
  \caption{GOP organization consisting of I-frames followed by predictive frames with MVs, and an HSV visualization of MVs, along with MB partitions.}
  \label{fig:codec}
\end{figure}

Codecs such as H.264 exploit spatial and temporal redundancy to enable efficient video storage and transmission. Rather than storing raw pixels, they suppress predictable information across frames, retaining motion information and residual signals sufficient for reconstruction. As shown in \cref{fig:codec}, video frames are categorized into three types: intra-coded (I), predictive (P), and bi-predictive (B). I-frames encode full spatial content as self-contained images. P-frames are predicted from a single past reference, while B-frames interpolate between past and future references. Frames are organized into Groups of Pictures (GoPs), each anchored by an I-frame followed by a sequence of P and B frames. Predictive frames are partitioned into macroblocks (MBs), each encoded via motion compensation: the encoder locates the most similar region in the reference frame and records the displacement as a MV. In H.264, block sizes range from $4{\times}4$ to $16{\times}16$ pixels, with smaller partitions used in regions exhibiting complex motion. The difference between the motion-compensated prediction and the true content is stored as a residual. 

Assembling MVs across a frame yields a displacement field that conceptually resembles optical flow. Unlike the latter, MVs are optimized for rate-distortion efficiency rather than photometric accuracy, resulting in lower spatial resolution and characteristic artifacts (\eg, noisy background motion and imprecise object boundaries). Despite these limitations, it provides a task-agnostic representation of scene motion available at zero marginal cost during bitstream parsing. 
Thus, the benefits of operating in the compressed domain for video analytics are threefold: (i) computational efficiency, as it bypasses full RGB decoding, while MB-level representations reduce dimensionality and accelerate downstream processing; (ii) inherent motion priors, which provide readily available temporal cues without requiring explicit optical flow computation; and (iii) privacy-preserving representations, which naturally abstract away sensitive pixel-level content (\eg, faces and license plates), thereby mitigating privacy risks.

\subsection{Detection and Tracking with Compressed Video Data}
The dominant line of research in early compressed-domain video analysis constructs per-MB motion activity maps from encoder-derived signals, including MVs, MB partition modes, quantization parameters, and residuals, followed by rule-based weighting and heuristic smoothing and thresholding to separate foreground moving objects from background \cite{Mezaris2004RealtimeCS, SOLANACIPRES200999, POPPE2009428, Patel2014H264MotionSegmentation, article_moura, 7478003}. Complementary approaches further reduce complexity by relying solely on MB-level coding decisions \cite{laumer2013compressed, laumer2016compressed}, underscoring the discriminative power of codec metadata alone. Temporal consistency is typically addressed via aggregation across consecutive frames or filtering incoherent vectors, while more structured methods employ spatio-temporal Markov Random Fields~\cite{bombardelli2018efficient}. Overall, these methods avoid RGB reconstruction but struggle with robustness in complex scenes and generalization.

With the rise of deep learning, compressed-domain signals are increasingly used as an auxiliary modality to improve efficiency. Most approaches follow a slow-fast paradigm, in which full RGB decoding and inference is performed only on sparse I-frames, while MVs and residual are leveraged to propagate features and detections across predictive frames. A first line of work uses hand-crafted MV-based aggregation at the bounding-box level for propagation \cite{Alvar2018MVYOLOMV, huang2025mvp, 9248145, jimaging9070132, 8575254}, remaining sensitive to noisy motion estimates and heuristic design choices. A second line instead learns propagation modules that better model temporal dynamics across predictive frames \cite{ duche2026see, article, 9010861, 10350828}. However, both approaches still rely on computationally expensive RGB inference in key-frames, and performance is often limited to the codec configuration seen during training. %Since our objective is to eliminate RGB processing rather than merely reduce its frequency, we do not consider such intermediate slow-fast designs as direct competitors. Likewise, approaches requiring explicitly computed optical flow are outside the comparison scope, since their additional computation is incompatible with the low-cost, CPU-oriented deployment target considered here.

CoVA~\cite{hwang2022cova} further reduces RGB usage by tracking coarse motion blobs. A lightweight temporal-aided UNet predicts motion objectness maps, which are converted to boxes via connected-component analysis and used to select a small set of frames for RGB decoding and trajectory refinement. This design, however, remains limited by MB-level precision, lacks semantic predictions at the blob stage, and struggles to separate overlapping objects because instances are defined by region connectivity. Moreover, its blob detector is trained per video to adapt to scene-specific compression statistics, limiting generalization.

\section{Methodology}
\subsection{Data Representation}
To emulate realistic edge-surveillance streams, we resize videos to 1080p and encode them at a target bitrate of 4~Mbps, restricting the minimum partition size to $8\times8$ pixels. We enforce stream causality by disabling bi-directional dependencies (B-frames), which is essential for low-latency inference without look-ahead buffering. The GoP size is set to the native frame rate, yielding one keyframe per second; this provides frequent error recovery under packet loss and limits long-term drift in accumulated MVs.

The compressed video frame is represented using a patch size of $p{=}16$ for maximum efficiency, producing a per-frame tensor of shape $\mathcal{X} \in \mathbb{R}^{\lceil H/p \rceil \times \lceil W/p \rceil \times 3}$. The first two channels store the horizontal and vertical displacements $(d_x, d_y)$, normalized by the dataset-wide 95th percentile of non-zero MV magnitudes and clipped to $[-2, 2]$. The third channel encodes the MB partition size $q \in \mathcal{M}$, where $\mathcal{M}=\{16{\times}16, \allowbreak 2{\times}(8{\times}16), \allowbreak 2{\times}(16{\times}8), \allowbreak 4{\times}(8{\times}8)\}$. For sub-partitioned MBs, constituent MVs are aggregated to the $16{\times}16$ patch by taking the maximum magnitude. To ensure spatial alignment with object locations at time $t$, MVs are remapped to their destination coordinates. 

I-frames are represented as zero tensors, which acts as a structured temporal dropout signal during training and forces the model to handle sparse but periodic motion gaps by extrapolating detections from preceding MV fields.

\subsection{MVDet}
\label{sec:mvdet_methodology}
\subsubsection{Architecture.} We adopt a lightweight, anchor-free detector based on CenterNet~\cite{zhou2019objects}. The architecture follows a encoder-decoder design with three predictions heads. Aggressive downsampling is omitted to preserve the already coarse spatial information. Channel widths and network depth are kept deliberately small for yielding a low computational cost. Detailed specifications are provided in \cref{tab:architecture}.

Partition modes $q$ are mapped into a continuous latent space via a learnable embedding table and concatenated channel-wise with the MVs before the first convolutional block. This allows the network to adaptively weight motion cues, such as emphasizing high-frequency motion in small partitions while suppressing background artifacts from bigger MBs.

After the first convolutional block, a recursive \textit{Temporal Gate} filters transient compression noise and jitter by propagating motion features across time. Unlike standard ConvGRU~\cite{convgru}, which uses three gated branches, our EMA-inspired update uses a single learned gate to interpolate between the previous state and current features. This yields roughly $3\times$ lower recurrent complexity, requiring only 0.5K parameters and 4.2 MFLOPs:
\begin{equation}
\begin{split}
g_t &= \sigma\bigl(\mathrm{Conv}_{1\times1}(\mathrm{Concat}(f_t, h_{t-1}))\bigr),\\
h_t &= \mathrm{GroupNorm}(g_t \odot h_{t-1} + (1 - g_t) \odot f_t)\, .
\end{split}
\end{equation}

Once robust temporal features are obtained, the encoder downsamples the grid only once using max pooling, followed by a second convolutional block. The bottleneck then applies three consecutive dilated convolutions to enlarge the receptive field while preserving the resolution for local-global motion consistency. The decoder restores the original feature grid via bilinear upsampling and a skip connection from the temporally gated features. Finally, three task-specific prediction heads with 1$\times$1 kernels produce a class-specific heatmap $\hat{Y}_c$ for center detection, a size map $\hat{S}$, and a local offset map $\hat{O}$ for sub-grid localization.

\begin{table}[tb]
\centering
\caption{MVDet architecture with tensor shapes for a single input frame. Each convolutional block (ConvBlock) consists of a convolution, batch normalization, and a LeakyReLU activation.}
\label{tab:architecture}
\begin{tabular}{@{}l@{\hspace{10pt}}l@{\hspace{10pt}}c@{}}
\toprule
Block & Layers & Output Shape \\
 \midrule
 Input &  MVs + MB Partition Embeddings & $(6,68,128)$ \\
 \midrule
 & ConvBlock & $(16,68,128)$ \\
 Encoder & Temporal Gate (S1)& $(16,68,128)$ \\
 & MaxPool & $(16,34,64)$ \\
 & ConvBlock & $(32,34,64)$ \\
 \midrule
Bottleneck & Dilated ConvBlocks ($d=\{2, 4, 8\}$) & $(32,34,64)$ \\
 \midrule
 & Bilinear Upsample & $(32,68,128)$ \\
 Decoder & Concat with S1 & $(48,68,128)$ \\
 & ConvBlock & $(16,68,128)$ \\
 \midrule
 & Heatmap & $(N_c,68,128)$ \\
 Heads & Size & $(2,68,128)$ \\
 & Offset & $(2,68,128)$ \\
\bottomrule
\end{tabular}
\end{table}
\subsubsection{Label creation.} Ground-truth labels follow a modified CenterNet paradigm. Each object is represented by an anisotropic Gaussian blob on a class-specific heatmap $Y_c$, with per-axis standard deviations proportional to the object width and height. Gaussian responses from overlapping objects are combined by an elementwise maximum operation, and the quantized center location is explicitly forced to attain a peak value of one. A $3\times3$ neighborhood around each quantized center defines the regression mask $R$. Within this region, a size map $S$ stores the object width and height, while an offset map $O$ encodes the displacement from each neighborhood cell to the true continuous center. 

When regression regions of multiple objects overlap, size and offset targets are assigned to the object whose center is closest to the corresponding cell, because that object is the one for which the cell provides the most local and least ambiguous center evidence. This tie-breaking rule is also consistent with the CenterNet formulation: at the detector output resolution, each cell, corresponding to a $16\times16$ pixel patch, can regress only a single object hypothesis. As a result, objects whose centers quantize to the same cell are inherently competing targets, and such collisions cannot be represented simultaneously by design. These four tensors, $\{Y_c, R, S, O\}$, constitute the dense supervision used by the detection head during training.

\subsubsection{Multi-task objective.} The model is trained with a weighted multi-task objective comprising a modified focal loss for heatmap prediction, $\ell_1$ losses for size and offset regression, and a Complete IoU (CIoU) loss~\cite{ciou_loss}. Following CenterNet, positive heatmap samples correspond to object-center peaks, while negative focal loss is evaluated only outside the $3\times3$ regression neighborhoods. Width, height, and offset regressions are supervised over the entire regression region using masked $\ell_1$ losses. In addition, predicted boxes decoded from the size and offset maps are refined using a CIoU loss evaluated at object centers to improve localization accuracy. The overall loss is:
\begin{equation}
\mathcal{L}
=
\lambda_{\mathrm{hm}}\mathcal{L}_{\mathrm{hm}}
+
\lambda_{\mathrm{wh}}\mathcal{L}_{\mathrm{wh}}
+
\lambda_{\mathrm{off}}\mathcal{L}_{\mathrm{off}}
+
\lambda_{\mathrm{ciou}}\mathcal{L}_{\mathrm{ciou}},
\end{equation}
with weights $\lambda_{\mathrm{hm}}=1.0$, $\lambda_{\mathrm{wh}}=0.1$, $\lambda_{\mathrm{off}}=1.0$, and $\lambda_{\mathrm{ciou}}=2.0$.

\subsubsection{Bounding box decoding.} At inference, boxes are decoded from ${\hat{Y}, \hat{S}, \hat{O}}$ using the same $3\times3$ neighborhood used for supervision. Rather than reading size and offset only at the peak, we compute a heatmap-weighted local expectation over the neighborhood, improving stability on the coarse MV grid. Candidate centers are obtained via non-maximum suppression on $\hat{Y}_c$ followed by top-$K$ selection, and each detection is assigned the class $\arg\max_c(\hat{Y}_c)$ at its center.

\subsection{MVLink}
\subsubsection{Stationary-object fragmentation.}
\label{sec:id_frag}
Compressed-domain moving-object tracking introduces an association failure mode that is uncommon in conventional TbD pipelines. When an object temporarily stops, it no longer induces a coherent MV pattern and can disappear from the detector output despite remaining visible in the scene. When motion resumes, the tracker must determine whether the new MV evidence belongs to an existing target or to a newly entering object. In the absence of appearance features, such stationary intervals are a major source of track fragmentation and identity switches.

Standard multi-object tracking (MOT) association strategies are not well suited to this setting because they typically handle all unmatched tracks uniformly, regardless of whether the missed detection is caused by occlusion, departure from the field of view, or a temporary stop. In Kalman-filter-based trackers, an unmatched track is usually marked as lost while its state continues to be propagated according to the last estimated velocity. For a stopped object, this prediction is actively harmful: the bounding box drifts away from the true location, reducing the likelihood of matching the object when it starts moving again.

\subsubsection{Perspective-invariant kinematic modeling.}
MVLink addresses this issue by minimally modifying the ByteTrack~\cite{zhang2022bytetrack} association pipeline: unmatched tracks are classified according to their recent motion state before propagation or termination. To reduce sensitivity to perspective-induced scale variation, MVLink computes kinematic features in a box-scale-normalized coordinate system. Given a Kalman filter state at frame $t$, the normalized speed and acceleration are defined as:
\begin{equation}
\begin{split}
v_{\text{norm}}^{(t)} = \sqrt{\frac{v_{x}^2 + v_{y}^2}{wh}}\, ,\\
a_{\text{norm}}^{(t)} = v_{\text{norm}}^{(t)} - v_{\text{norm}}^{(t-1)}\, ,
\end{split}
\end{equation}
where $v_x$ and $v_y$ are the image-plane velocity components, and $w$ and $h$ are the current box width and height. MVLink maintains exponential moving averages of both quantities to suppress frame-level noise and obtain stable estimates. When a track is not matched to an MV detection, MVLink inspects the smoothed kinematic history. If the normalized speed falls below a stationary threshold and the object was decelerating, the track is assigned to a dedicated \textit{Stopped} state; otherwise, it is marked as \textit{Lost}. For \textit{Stopped} tracks, MVLink keeps the boxes anchored at their last reliable location for a fixed temporal window. As a result, when objects resume motion and again produce MV detections, the new observations can be matched to the preserved tracks rather than initializing new identities.

\section{Experiments and Results}
\subsection{Dataset}
We train and evaluate MVTrack on the VIRAT Ground Dataset~\cite{virat}, specifically the DIVA-V1 subset~\cite{divaV1}, a large-scale surveillance benchmark recorded from static overhead CCTV cameras across 5 different outdoor scenes including parking lots, walkways, and building entrances. It comprises 119 videos totaling 4.3 hours at a frame rate of 30 FPS, with 55 videos reserved for validation.

DIVA-V1 provides dense annotations for activity recognition and MOT, spanning pedestrians, vehicles, and other object categories. We leverage these complementary annotations to partition object trajectories into dynamic track segments, retaining only the intervals during which an object participates in an annotated non-static activity. This excludes stationary periods, such as pedestrians sitting or crouching and parked vehicles, while tolerating minor boundary noise of a few frames around static-dynamic transitions.

\subsection{Metrics and Evaluation}
\label{subsec:metrics_eval}
We report $\text{mAP}_{50:95}$~\cite{lin2014microsoft} as the primary detection metric, together with per-class scores and $\text{mAP}_{50}$~\cite{map50}, which characterizes the trade-off between recall and bounding-box precision under coarse input resolution. Tracking performance is primarily assessed with HOTA~\cite{hota}, which jointly captures detection accuracy (DetA) and association quality (AssA), while MOTA~\cite{mota} and IDF1~\cite{Ristani2016PerformanceMA} are included for completeness.

To ensure a fair comparison between conventional MOT trackers (\ie, tracking all objects regardless of motion state) and MVTrack, we adopt a dedicated evaluation protocol. Rather than removing static track segments from the ground truth, which would unfairly penalize conventional trackers via false positives, we mark them as distractors, thereby leveling the detection benchmark. At the same time, this protocol preserves the association challenge discussed in \cref{sec:id_frag}: conventional RGB trackers can observe stopped objects and maintain continuous identities, whereas MVTrack must re-identificate after stationary periods, making association intrinsically harder for MVTrack.

\subsection{Implementation details}
MVDet is trained to detect moving objects for 100 epochs with a batch size of 64, utilizing a temporal window of $T=10$ for backpropagation through time (BPTT). This horizon is strictly confined to bounding gradient history during training; at inference, the recursive formulation processes frames sequentially and fully continuously, meaning the training window size has zero impact on inference latency or memory footprint. We employ the Adam\cite{kingma2015adam} optimizer with an initial learning rate of $6 \times 10^{-3}$, which is halved upon mAP stagnation. Data augmentation includes random translations and horizontal flips.

The recently introduced YOLO26n \cite{yolo26_ultralytics} is adopted as the sole RGB baseline because it combines state-of-the-art performance with a CPU-efficient design tailored for edge deployment, providing a strong test of whether compressed-domain motion can outperform an RGB-only detector under practical CPU constraints. Accordingly, we use its standard $640\times640$ input resolution, avoiding the substantial computational overhead associated with higher resolutions. The model is trained to detect moving and static targets for 100 epochs with a batch size of 64 using the default hyperparameter configuration.

Hyperparameters for the different association algorithms are tuned using the Optuna framework \cite{optuna_2019}, targeting the most crowded sequences per-scene within the training split to ensure robustness in complex scenarios.

\subsection{Ablations}
\subsubsection{Architecture.} We ablate the key design components of MVDet, beginning with temporal modeling as it is expected to have the largest impact in MV-based detection. For this analysis, MVs are used as the sole input modality. As shown in \cref{tab:ablation_arch}, the single-frame \textit{Baseline} achieves the lowest accuracy. Conventional many-to-one aggregation schemes (\textit{Mean Pooling} and \textit{Channel Stacking}) improve robustness to transient artifacts, but discard temporal ordering and operate over a fixed temporal windows. In contrast, the proposed \textit{Temporal Gate} selectively accumulates motion evidence over time, consistently outperforming both windowed alternatives. Increasing the BPTT horizon yields substantial gains up to $T=5$, after which improvements saturate; we therefore adopt $T=10$ as the default configuration. Adding MB partition metadata further improves performance for both classes with negligible overhead, indicating that codec block structure provides complementary spatial cues beyond the MV field alone.
\begin{table}[tb]
  \caption{Comparison of temporal aggregation strategies and the injection of MB partitions on detection accuracy.}
  \label{tab:ablation_arch}
  \centering
  \begin{tabular}{@{}l@{\hspace{10pt}}cccc@{}}
    \toprule
    Architecture & mAP & mAP$_{50}$ & mAP$_{p}$ & mAP$_{v}$\\
    \midrule
    Baseline ($T=1$)  & 28.05 & 59.79 & 13.56 & 42.54 \\
    Mean ($T=5$) & 31.85 & 67.73 & 17.44 & 46.26\\
    Channel Stack ($T=5$) & 32.04 & 67.54 & 17.68 & 46.40\\
    Temporal Gate ($T=3$) & 32.40 & 68.66 & 18.02 & 46.78\\
    Temporal Gate ($T=5$) & 34.51 & 71.49 & 20.37 & 48.65 \\
    Temporal Gate ($T=10$) & 34.79 & 72.33 & 20.67 & 48.92\\
    \midrule
    + MB Partition & 36.45 & 73.77 & 22.42 & 50.47\\
  \bottomrule
  \end{tabular}
\end{table}

\subsubsection{Scene generalization.} The dataset comprises only five scenes that appear in both splits, limiting the assessment of cross-scene generalization. We therefore perform Leave-One-Scene-Out (LOSO) cross-validation, training on four scenes and evaluating on the held-out fifth. As shown in~\cref{tab:ablation_loso}, MVDet achieves an average mAP of 31.91, remaining within 5 points of the standard setting. Performance remains stable across LOSO folds, suggesting that MVDet primarily exploits transferable motion patterns, with only a modest contribution from scene-specific statistics.
\begin{table}[tb]
  \caption{Cross-scene validation to evaluate spatial domain invariance across distinct sequences.}
  \label{tab:ablation_loso}
  \centering
  \begin{tabular}{@{}c@{\hspace{10pt}}cccc@{}}
    \toprule
    Held-out Seq. & mAP & mAP$_{50}$ & mAP$_{p}$ & mAP$_{v}$\\
    \midrule
    0000 & 25.37 & 52.14 & 11.83 & 38.91 \\
    0002 & 31.08 & 65.12 & 20.94 & 41.22 \\
    0400 & 38.84 & 72.91 & 24.67 & 53.01 \\
    0401 & 27.55 & 60.41 & 16.73 & 38.37 \\
    0500 & 36.72 & 71.22 & 27.65 & 45.79 \\
    \midrule
    LOSO & 31.91 & 64.36 & 20.36 & 43.46\\
    \bottomrule
  \end{tabular}
\end{table}

\subsubsection{Encoding generalization.} 
Compressed-domain models must be robust to codec settings that shift the MV distribution, including temporal sampling density, predictive-frame chain length, MV quantization noise, and bi-predictive coding. \Cref{tab:ablation_encoding} shows that MVDet, trained with the baseline encoding, remains stable under these changes. The largest degradation occurs at lower frame rates, particularly for vehicles, whose larger inter-frame displacements move farther outside the baseline training distribution. In contrast, doubling the GoP length has little effect, and performance remains strong even at a target bitrate of 1 Mbps, a regime representative of budget low-latency IP cameras. We further evaluate a setting allowing the use B-frames. Future-referenced MVs are mapped to the forward convention used by P-frames by reversing their sign, which approximates motion under near-constant velocity between adjacent frames. 
\begin{table}[tb]
  \caption{Model robustness under variable codec settings and bitstream degradations.}
  \label{tab:ablation_encoding}
  \centering
  \begin{tabular}{@{}l@{\hspace{10pt}}cccc@{}}
    \toprule
    Codec Config. Variation & mAP & mAP$_{50}$ & mAP$_{p}$ & mAP$_{v}$\\
    \midrule
    FPS 25 & 36.29 & 74.29 & 22.86 & 49.73 \\
    FPS 20 & 35.08 & 73.49 & 21.91 & 48.26 \\
    FPS 15 & 33.73 & 73.36 & 21.50 & 45.95 \\
    GOP 2 s & 36.35 & 73.29 & 22.05 & 50.56 \\
    Bitrate 2 Mbps & 35.36 & 73.82 & 21.93 & 48.79 \\
    Bitrate 1 Mbps & 34.93 & 73.12 & 21.60 & 48.27 \\
    B-frames usage & 35.01 & 72.54 & 21.07 & 48.96 \\
    \bottomrule
  \end{tabular}
\end{table}

\subsection{RGB Tracking Comparison}
To benchmark against conventional RGB-based MOT pipelines, we first pair both MVDet and YOLO26n with ByteTrack under the dynamic evaluation protocol. As shown in \cref{tab:rgb_comparison}, MVDet with ByteTrack reaches a HOTA score close to the RGB baseline. It also obtains higher DetA and MOTA, suggesting that MV inputs provide strong evidence for detecting active objects. This advantage is especially pronounced for pedestrians, since MVDet operates on MVs extracted from the native resolution, whereas YOLO26n processes resized RGB frames that can suppress small-object detail.

The main limitation of the ByteTrack variant is association quality: its lower AssA reflects the difficulty of maintaining identities when motion-only detections are temporarily unavailable. MVLink directly targets this failure mode. By spotting stopped tracks through kinematic cues, it substantially improves AssA, IDF1, and HOTA, surpassing the RGB baseline on the latter two metrics. The residual AssA and vehicle-HOTA gaps are likely caused by long stationary intervals, especially parked vehicles, that exceed MVLink's stopped-track horizon. Thus, MVTrack recovers much of the identity continuity lost in a purely motion-driven setting while retaining the efficiency and privacy benefits of compressed-domain tracking.
\begin{table}[tb]
  \caption{Tracking performance under the dynamic evaluation protocol.}
  \label{tab:rgb_comparison}
  \centering
  \begin{tabular}{@{}lc@{\hspace{10pt}}ccccccc@{}}
    \toprule
    Detector & Association Module & HOTA & HOTA$_p$ & HOTA$_v$ & DetA & AssA & MOTA & IDF1\\
    \midrule
    YOLO26n & ByteTrack & 52.78 & 41.00 & \textbf{64.56} & 48.97 & \textbf{57.24} & 54.56 & 66.94\\
    MVDet & ByteTrack & 51.92 & 44.48 & 59.36 & 53.34 & 50.91 & 62.18 & 65.86\\
    \midrule
    MVDet & MVLink & \textbf{53.88} & \textbf{45.81} & 61.95 & \textbf{53.47} & 54.63 & \textbf{63.68} & \textbf{70.50} \\
    \bottomrule
  \end{tabular}
\end{table}

As reported in \cref{tab:rgb_comparison_cost}, MVDet results are achieved with over 60$\times$ fewer parameters and 40$\times$ fewer FLOPs, while also reducing inference latency by nearly one order of magnitude on a consumer CPU. We report detector latency (forward pass, bounding-box decoding, and post-processing), as it dominates runtime in TbD pipelines without appearance-based re-identification. Consequently, we do not separately report the overhead of bitstream parsing compared to full RGB decoding, or data association, as these components are detector-agnostic and contribute only a minor fraction of the end-to-end runtime. When exported to ONNX, MVDet further reduces latency to 2.17 ms, enabling real-time processing of multiple streams on edge devices.
\begin{table}[tb]
  \caption{Computational cost comparison of MVDet and YOLO26n.}
  \label{tab:rgb_comparison_cost}
  \centering
  \begin{tabular}{@{}l@{\hspace{10pt}}ccc@{}}
    \toprule
    Detector & Params & FLOPs & Latency (ms)\\
    \midrule
    YOLO26n & 2.5M & 5.4B & 44.80\\
    MVDet & 41K & 139M & 5.21\\
    \bottomrule
  \end{tabular}
\end{table}

\subsection{Qualitative Results} 
\begin{figure}[tb]
  \centering
  \begin{subfigure}{\linewidth}
    \includegraphics[width=\linewidth]{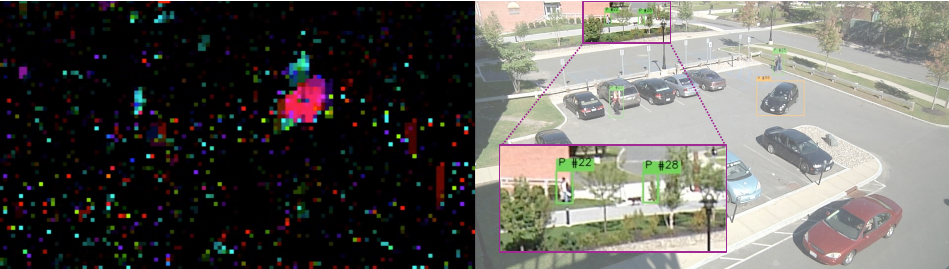}
    \caption{}
    \label{fig:qual1}
  \end{subfigure}
  \vspace{5pt}
  \begin{subfigure}{\linewidth}
    \includegraphics[width=\linewidth]{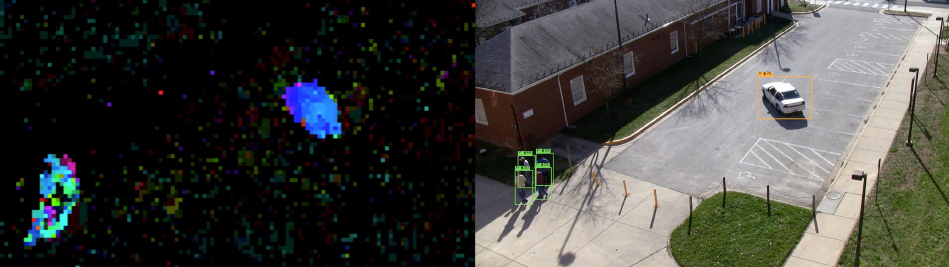}
    \caption{}
    \label{fig:qual2}
  \end{subfigure}
  \vspace{5pt}
  \begin{subfigure}{\linewidth}
    \includegraphics[width=\linewidth]{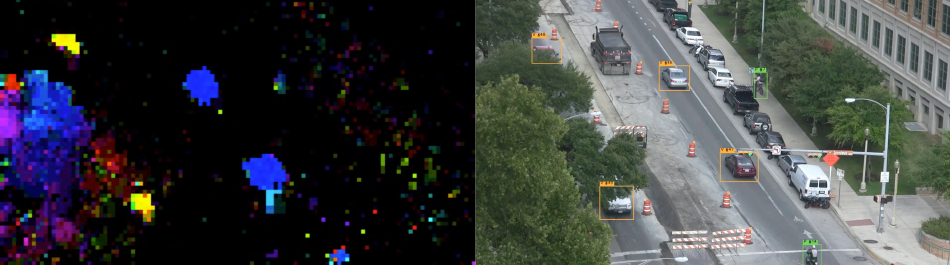}
    \caption{}
    \label{fig:qual3}
  \end{subfigure}
  \caption{Qualitative results of MVTrack under challenging surveillance conditions.}
  \label{fig:qualpos}
\end{figure}

\Cref{fig:qual1} highlights MVDet's ability to detect small moving objects. Even distant pedestrians that occupy only a few pixels produce coherent motion patterns that are sufficient for localization. The example also illustrates the benefit of aggregating motion over multiple frames: detections remain stable through short partial occlusions, such as the tree occlusion shown in the sequence.

\Cref{fig:qual2} shows a scene with substantial pedestrian overlap and strong illumination variation, making instance separation challenging from RGB appearance alone. In contrast, the HSV visualization of the MV field reveals a coherent motion structure for the group, from which MVDet recovers individual bounding boxes. At the same time, it effectively disentangles pedestrians from their cast shadows.

\Cref{fig:qual1} and \cref{fig:qual3} further demonstrate the task-specific selectivity of the proposed representation. Moving distractors such as dogs are ignored, while static scene elements such as parked vehicles do not trigger detections. In \cref{fig:qual3}, cyclists and motorcyclists are localized and assigned to the pedestrian class despite their larger MV magnitudes. Across these examples, MVDet remains robust to dynamic background motion, such as swaying trees, and to visible compression artifacts, supporting MVs as a reliable signal for surveillance tracking.

\begin{figure}[tb]
  \centering
  \begin{subfigure}[t]{0.25\linewidth}
    \includegraphics[width=\linewidth]{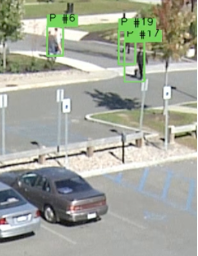}
    \caption{}
    \label{fig:fail1}
  \end{subfigure}
  \hfill
  \begin{subfigure}[t]{0.27\linewidth}
    \includegraphics[width=\linewidth]{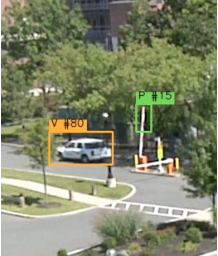}
    \caption{}
    \label{fig:fail2}
  \end{subfigure}
  \hfill
  \begin{subfigure}[t]{0.25\linewidth}
    \includegraphics[width=\linewidth]{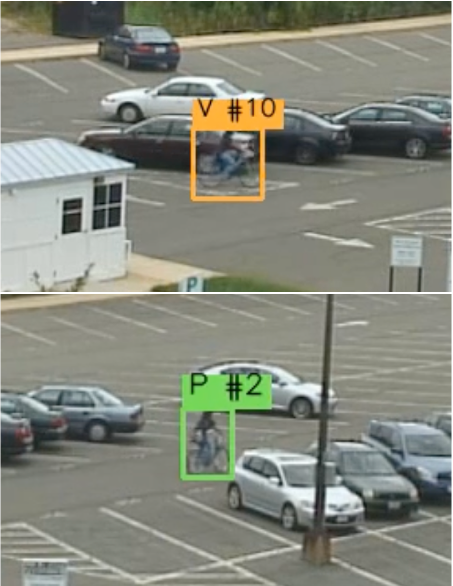}
    \caption{}
    \label{fig:fail3}
  \end{subfigure}
  \caption{Representative failure cases of MVTrack.}
  \label{fig:qualfail}
\end{figure}

The failure cases in \cref{fig:qualfail} illustrate the main limitations of relying exclusively on compressed-domain motion. Although MVDet handles large overlapping objects and small isolated pedestrians well, performance degrades when both conditions occur simultaneously. Small pedestrians moving in close proximity can generate highly correlated MV patterns that collapse into a single motion blob. As shown in \cref{fig:fail1}, this causes two neighboring pedestrians to be merged into one detection. This behavior is further amplified by MVDet's architecture, as discussed in \cref{sec:mvdet_methodology}, because each output cell can regress only a single bounding box. The example also highlights a fundamental trade-off of compressed-domain analysis: even when the object is detected, localization can be imprecise because MV fields are spatially coarse and affected by codec quantization.

Ambiguous motion from non-target objects provides a second source of errors. In \cref{fig:fail2}, a rigid moving object produces a motion pattern similar to that of a pedestrian, resulting in a false positive. We also observe a mild sensitivity to motion-blob geometry: vertically elongated blobs can favor pedestrian predictions, whereas horizontally elongated blobs can favor vehicle predictions. As shown in \cref{fig:fail3}, this can cause a cyclist to be initially classified as a vehicle before being correctly reassigned once the observed motion structure changes. Nevertheless, this effect is limited in aggregate, as MVDet still achieves higher overall detection metrics than the RGB-based detector.

\section{Summary and Future Work}
We presented MVTrack, a tracking framework for moving objects directly from compressed H.264 bitstreams. Instead of using MVs as an auxiliary shortcut between periodic RGB detections, MVTrack employs an always-fast design where codec metadata serves as the primary sensing signal. The framework combines MVDet, a compact CenterNet-style detector with gated temporal aggregation and MB-partition cues, with MVLink, a modified ByteTrack association module that preserves identities through short stationary periods. On VIRAT, MVTrack surpasses the RGB YOLO26n-ByteTrack baseline while using substantially fewer parameters and FLOPs, and achieving lower CPU latency. The experiments also clarify why compressed-domain tracking is effective. Temporal aggregation improves robustness to noisy MV fields, while MB-partition metadata provides useful spatial structure. Overall, MVDet learns transferable motion patterns rather than merely memorizing scene-specific statistics, and remains effective beyond a single controlled compression setting, including challenging cases with small objects or severe overlaps.

More broadly, MVTrack suggests a practical role for compressed-domain models as always-on front ends for large-scale video analytics. A lightweight MV tracker can continuously monitor many streams on CPU-only edge hardware, skip static intervals, and trigger selective RGB decoding or higher-capacity models only when meaningful activity occurs. Future work should evaluate the transferability of MVTrack to infrared imagery, severe weather and precipitation, additional codecs such as H.265, and a wider range of camera viewpoints and deployment conditions. Another promising direction is to extend the framework toward conventional MOT by initializing tracks with an RGB detector at the beginning of a sequence, thereby exposing objects that are initially stationary before relying primarily on compressed-domain motion. Although our primary focus was on detection, MV fields could also be exploited more explicitly during association, for example through velocity-aware affinity terms or by replacing Kalman filters.

Overall, these findings indicate that MVs are not merely a computational shortcut for RGB pipelines, but a powerful representation for scalable, privacy-preserving, and resource-efficient surveillance tracking.

\section*{Acknowledgements}
This work has been partially supported by the Spanish project PID2025-170489NB-I00, by ICREA under the ICREA Academia programme, and by the Milestone Research Program at the University of Barcelona.

\bibliographystyle{splncs04}
\bibliography{main}

@String(CVPR  = {IEEE Conf. Comput. Vis. Pattern Recog.})

@String(ICCV  = {Int. Conf. Comput. Vis.})

@String(ECCV  = {Eur. Conf. Comput. Vis.})

@String(ICLR  = {Int. Conf. Learn. Represent.})

@String(CVPRW = {IEEE Conf. Comput. Vis. Pattern Recog. Worksh.})

@String(CVPR  = {CVPR})

@String(ICCV  = {ICCV})

@String(ECCV  = {ECCV})

@String(ICLR  = {ICLR})

@String(CVPRW = {CVPRW})

@INPROCEEDINGS{8397016,
  author={Ranjbar Alvar, Saeed and Choi, Hyomin and Bajic, Ivan V},
  booktitle={2018 IEEE Conference on Multimedia Information Processing and Retrieval (MIPR)}, 
  title={Can You Tell a Face from a HEVC Bitstream?}, 
  year={2018},
  volume={},
  number={},
  pages={257-261},
  doi={10.1109/MIPR.2018.00060}}

@INPROCEEDINGS{9010861,
  author={Wang, Shiyao and Group, Alibaba and Lu, Hongchao and Deng, Zhidong},
  booktitle={2019 IEEE/CVF International Conference on Computer Vision (ICCV)}, 
  title={Fast Object Detection in Compressed Video}, 
  year={2019},
  volume={},
  number={},
  pages={7103-7112},
  doi={10.1109/ICCV.2019.00720}}

@INPROCEEDINGS{10350828,
  author={Tran, Ryan and Kanaujia, Atul and Parameswaran, Vasu},
  booktitle={2023 IEEE/CVF International Conference on Computer Vision Workshops (ICCVW)}, 
  title={Fast Object Detection in High-Resolution Videos}, 
  year={2023},
  volume={},
  number={},
  pages={1461-1470},
  doi={10.1109/ICCVW60793.2023.00159}}

@article{article,
author = {Liu, Qiankun and Liu, Bin and Wu, Yue and Li, Weihai and Yu, Nenghai},
year = {2019},
month = {01},
pages = {76489-76499},
title = {Real-Time Online Multi-Object Tracking in Compressed Domain},
volume = {7},
journal = {IEEE Access},
doi = {10.1109/ACCESS.2019.2921975}
}

@Article{jimaging9070132,
AUTHOR = {True, Julian and Khan, Naimul},
TITLE = {Motion Vector Extrapolation for Video Object Detection},
JOURNAL = {Journal of Imaging},
VOLUME = {9},
YEAR = {2023},
NUMBER = {7},
ARTICLE-NUMBER = {132},
PubMedID = {37504809},
ISSN = {2313-433X},
DOI = {10.3390/jimaging9070132}
}

@article{Alvar2018MVYOLOMV,
  title={MV-YOLO: Motion Vector-Aided Tracking by Semantic Object Detection},
  author={Saeed Ranjbar Alvar and Ivan V. Baji{\'c}},
  journal={2018 IEEE 20th International Workshop on Multimedia Signal Processing (MMSP)},
  year={2018},
  pages={1-5},
  url={https://api.semanticscholar.org/CorpusID:25756869}
}

@INPROCEEDINGS{9248145,
  author={Bommes, Lukas and Lin, Xinlin and Zhou, Junhong},
  booktitle={2020 15th IEEE Conference on Industrial Electronics and Applications (ICIEA)}, 
  title={MVmed: Fast Multi-Object Tracking in the Compressed Domain}, 
  year={2020},
  volume={},
  number={},
  pages={1419-1424},
  doi={10.1109/ICIEA48937.2020.9248145}}

@inproceedings{hwang2022cova,
  title={CoVA: Exploiting Compressed-Domain analysis to accelerate video analytics},
  author={Hwang, Jinwoo and Kim, Minsu and Kim, Daeun and Nam, Seungho and Kim, Yoonsung and Kim, Dohee and Sharma, Hardik and Park, Jongse},
  booktitle={2022 USENIX annual technical conference (USENIX ATC 22)},
  pages={707--722},
  year={2022}
}

@article{article_moura,
author = {Moura, Ronaldo and Hemerly, Elder and Cunha, Adilson},
year = {2014},
month = {05},
pages = {},
title = {Temporal Motion Vector Filter for Fast Object Detection on Compressed Video},
volume = {29},
journal = {JOURNAL OF COMMUNICATIONS AND INFORMATION SYSTEMS},
doi = {10.14209/jcis.2014.1}
}

@article{Nguyen2020TowardSV,
  title={Toward Scalable Video Analytics Using Compressed-Domain Features at the Edge},
  author={Dien Nguyen and Jaehyuk Choi},
  journal={Applied Sciences},
  year={2020},
  url={https://api.semanticscholar.org/CorpusID:226733466}
}

@INPROCEEDINGS{8575254,
  author={Ujiie, Takayuki and Hiromoto, Masayuki and Sato, Takashi},
  booktitle={2018 IEEE/CVF Conference on Computer Vision and Pattern Recognition Workshops (CVPRW)}, 
  title={Interpolation-Based Object Detection Using Motion Vectors for Embedded Real-time Tracking Systems}, 
  year={2018},
  volume={},
  number={},
  pages={729-7298},
  doi={10.1109/CVPRW.2018.00104}}

@INPROCEEDINGS{9251974,
  author={Song, Zhuoran and Wu, Feiyang and Liu, Xueyuan and Ke, Jing and Jing, Naifeng and Liang, Xiaoyao},
  booktitle={2020 53rd Annual IEEE/ACM International Symposium on Microarchitecture (MICRO)}, 
  title={VR-DANN: Real-Time Video Recognition via Decoder-Assisted Neural Network Acceleration}, 
  year={2020},
  volume={},
  number={},
  pages={698-710},
  doi={10.1109/MICRO50266.2020.00063}}

@INPROCEEDINGS{9879197,
  author={Xu, Kai and Yao, Angela},
  booktitle={2022 IEEE/CVF Conference on Computer Vision and Pattern Recognition (CVPR)}, 
  title={Accelerating Video Object Segmentation with Compressed Video}, 
  year={2022},
  volume={},
  number={},
  pages={1332-1341},
  doi={10.1109/CVPR52688.2022.00140}}

@ARTICLE{8982035,
  author={Tan, Zhentao and Liu, Bin and Chu, Qi and Zhong, Hangshi and Wu, Yue and Li, Weihai and Yu, Nenghai},
  journal={IEEE Transactions on Circuits and Systems for Video Technology}, 
  title={Real Time Video Object Segmentation in Compressed Domain}, 
  year={2021},
  volume={31},
  number={1},
  pages={175-188},
  doi={10.1109/TCSVT.2020.2971641}}

@ARTICLE{1294953,
  author={Mezaris, V. and Kompatsiaris, I. and Boulgouris, N.V. and Strintzis, M.G.},
  journal={IEEE Transactions on Circuits and Systems for Video Technology}, 
  title={Real-time compressed-domain spatiotemporal segmentation and ontologies for video indexing and retrieval}, 
  year={2004},
  volume={14},
  number={5},
  pages={606-621},
  doi={10.1109/TCSVT.2004.826768}}

@InProceedings{Hu_2023_CVPR,
    author    = {Hu, Yubin and He, Yuze and Li, Yanghao and Li, Jisheng and Han, Yuxing and Wen, Jiangtao and Liu, Yong-Jin},
    title     = {Efficient Semantic Segmentation by Altering Resolutions for Compressed Videos},
    booktitle = {Proceedings of the IEEE/CVF Conference on Computer Vision and Pattern Recognition (CVPR)},
    month     = {June},
    year      = {2023},
    pages     = {22627-22637}
}

@INPROCEEDINGS{11259157,
  author={Huang, Binhua and Wang, Ni and Pakrashi, Arjun and Dev, Soumyabrata},
  booktitle={2025 18th International Congress on Image and Signal Processing, BioMedical Engineering and Informatics (CISP-BMEI)}, 
  title={MoCLIP-Lite: Efficient Video Recognition by Fusing CLIP with Motion Vectors}, 
  year={2025},
  volume={},
  number={},
  pages={1-6},
  doi={10.1109/CISP-BMEI68103.2025.11259157}}

@inproceedings{wu2018compressed,
  title={Compressed video action recognition},
  author={Wu, Chao-Yuan and Zaheer, Manzil and Hu, Hexiang and Manmatha, R and Smola, Alexander J and Kr{\"a}henb{\"u}hl, Philipp},
  booktitle={Proceedings of the IEEE conference on computer vision and pattern recognition (CVPR)},
  pages={6026--6035},
  year={2018}
}

@INPROCEEDINGS{7780666,
  author={Zhang, Bowen and Wang, Limin and Wang, Zhe and Qiao, Yu and Wang, Hanli},
  booktitle={2016 IEEE Conference on Computer Vision and Pattern Recognition (CVPR)}, 
  title={Real-Time Action Recognition with Enhanced Motion Vector CNNs}, 
  year={2016},
  volume={},
  number={},
  pages={2718-2726},
  doi={10.1109/CVPR.2016.297}}

@article{gronquist2023efficient,
  title={Efficient temporally-aware deepfake detection using h. 264 motion vectors},
  author={Gr{\"o}nquist, Peter and Ren, Yufan and He, Qingyi and Verardo, Alessio and S{\"u}sstrunk, Sabine},
  journal={arXiv preprint arXiv:2311.10788},
  year={2023}
}

@article{POPPE2009428,
title = {Moving object detection in the H.264/AVC compressed domain for video surveillance applications},
journal = {Journal of Visual Communication and Image Representation},
volume = {20},
number = {6},
pages = {428-437},
year = {2009},
issn = {1047-3203},
doi = {https://doi.org/10.1016/j.jvcir.2009.05.001},
url = {https://www.sciencedirect.com/science/article/pii/S1047320309000650},
author = {Chris Poppe and Sarah {De Bruyne} and Tom Paridaens and Peter Lambert and Rik {Van de Walle}},
}

@article{SOLANACIPRES200999,
title = {Real-time moving object segmentation in H.264 compressed domain based on approximate reasoning},
journal = {International Journal of Approximate Reasoning},
volume = {51},
number = {1},
pages = {99-114},
year = {2009},
issn = {0888-613X},
doi = {https://doi.org/10.1016/j.ijar.2009.09.002},
url = {https://www.sciencedirect.com/science/article/pii/S0888613X0900139X},
author = {C. Solana-Cipres and G. Fernandez-Escribano and L. Rodriguez-Benitez and J. Moreno-Garcia and L. Jimenez-Linares},
}

@ARTICLE{7478003,
  author={Chen, Yung-Wei and Chen, Kai and Yuan, Shih-Yi and Kuo, Sy-Yen},
  journal={IEEE Access}, 
  title={Moving Object Counting Using a Tripwire in H.265/HEVC Bitstreams for Video Surveillance}, 
  year={2016},
  volume={4},
  number={},
  pages={2529-2541},
  doi={10.1109/ACCESS.2016.2572121}}

@inproceedings{zhou2023mvflow,
  title={MVFlow: deep optical flow estimation of compressed videos with motion vector prior},
  author={Zhou, Shili and Jiang, Xuhao and Tan, Weimin and He, Ruian and Yan, Bo},
  booktitle={Proceedings of the 31st ACM International Conference on Multimedia},
  pages={1964--1974},
  year={2023}
}

@INPROCEEDINGS{11095001,
  author={Zhao, Zijia and Huo, Yuqi and Yue, Tongtian and Guo, Longteng and Lu, Haoyu and Wang, Bingning and Chen, Weipeng and Liu, Jing},
  booktitle={2025 IEEE/CVF Conference on Computer Vision and Pattern Recognition (CVPR)}, 
  title={Efficient Motion-Aware Video MLLM}, 
  year={2025},
  volume={},
  number={},
  pages={24159-24168},
  doi={10.1109/CVPR52734.2025.02250}}

@article{sarkar2026cope,
  title={CoPE-VideoLM: Leveraging Codec Primitives For Efficient Video Language Modeling},
  author={Sarkar, Sayan Deb and Pautrat, R{\'e}mi and Miksik, Ondrej and Pollefeys, Marc and Armeni, Iro and Rad, Mahdi and Dusmanu, Mihai},
  journal={arXiv preprint arXiv:2602.13191},
  year={2026}
}

@inproceedings{10.1145/3394171.3413504,
author = {Chen, Peilin and Yang, Wenhan and Sun, Long and Wang, Shiqi},
title = {When Bitstream Prior Meets Deep Prior: Compressed Video Super-resolution with Learning from Decoding},
year = {2020},
isbn = {9781450379885},
publisher = {Association for Computing Machinery},
address = {New York, NY, USA},
url = {https://doi.org/10.1145/3394171.3413504},
doi = {10.1145/3394171.3413504},
booktitle = {Proceedings of the 28th ACM International Conference on Multimedia},
pages = {1000–1008},
numpages = {9},
location = {Seattle, WA, USA},
series = {MM '20}
}

@ARTICLE{9509352,
  author={Chen, Peilin and Yang, Wenhan and Wang, Meng and Sun, Long and Hu, Kangkang and Wang, Shiqi},
  journal={IEEE Transactions on Image Processing}, 
  title={Compressed Domain Deep Video Super-Resolution}, 
  year={2021},
  volume={30},
  number={},
  pages={7156-7169},
  doi={10.1109/TIP.2021.3101826}}

@ARTICLE{7742914,
  author={Xu, Mai and Jiang, Lai and Sun, Xiaoyan and Ye, Zhaoting and Wang, Zulin},
  journal={IEEE Transactions on Image Processing}, 
  title={Learning to Detect Video Saliency With HEVC Features}, 
  year={2017},
  volume={26},
  number={1},
  pages={369-385},
  doi={10.1109/TIP.2016.2628583}}

@article{huang2025mvp,
  title={MVP: Motion Vector Propagation for Zero-Shot Video Object Detection},
  author={Huang, Binhua and Wang, Ni and Yao, Wendong and Dev, Soumyabrata},
  journal={arXiv preprint arXiv:2509.18388},
  year={2025}
}

@inproceedings{zhou2019objects,
  title={Objects as Points},
  author={Zhou, Xingyi and Wang, Dequan and Kr{\"a}henb{\"u}hl, Philipp},
  booktitle={arXiv preprint arXiv:1904.07850},
  year={2019}
}

@misc{yolo26_ultralytics,
  author = {Glenn Jocher and Jing Qiu},
  title = {Ultralytics {YOLO26}},
  version = {26.0.0},
  year = {2026},
  url = {https://github.com/ultralytics/ultralytics},
  orcid = {0000-0001-5950-6979, 0000-0003-3783-7069},
  license = {AGPL-3.0}
}

@INPROCEEDINGS{virat,
  author={Oh, Sangmin and Hoogs, Anthony and Perera, Amitha and Cuntoor, Naresh and Chen, Chia-Chih and Lee, Jong Taek and Mukherjee, Saurajit and Aggarwal, J. K. and Lee, Hyungtae and Davis, Larry and Swears, Eran and Wang, Xioyang and Ji, Qiang and Reddy, Kishore and Shah, Mubarak and Vondrick, Carl and Pirsiavash, Hamed and Ramanan, Deva and Yuen, Jenny and Torralba, Antonio and Song, Bi and Fong, Anesco and Roy-Chowdhury, Amit and Desai, Mita},
  booktitle={CVPR 2011}, 
  title={A large-scale benchmark dataset for event recognition in surveillance video}, 
  year={2011},
  volume={},
  number={},
  pages={3153-3160},
  doi={10.1109/CVPR.2011.5995586}}

@ARTICLE{ciou_loss,
  author={Zheng, Zhaohui and Wang, Ping and Ren, Dongwei and Liu, Wei and Ye, Rongguang and Hu, Qinghua and Zuo, Wangmeng},
  journal={IEEE Transactions on Cybernetics}, 
  title={Enhancing Geometric Factors in Model Learning and Inference for Object Detection and Instance Segmentation}, 
  year={2022},
  volume={52},
  number={8},
  pages={8574-8586},
  doi={10.1109/TCYB.2021.3095305}}

@techreport{divaV1,
  title     = {Data, Algorithms, and Framework for Automated Analytics of Videos with Activities},
  author    = {Kitware, Inc.},
  institution = {Intelligence Advanced Research Projects Activity (IARPA), DIVA Program},
  year      = {2019},
  note      = {{DIVA-V1} dataset. Annotations publicly released via NIST ActEV. Available: \url{https://viratdata.org/}}
}

@InProceedings{lin2014microsoft,
author="Lin, Tsung-Yi
and Maire, Michael
and Belongie, Serge
and Hays, James
and Perona, Pietro
and Ramanan, Deva
and Doll{\'a}r, Piotr
and Zitnick, C. Lawrence",
editor="Fleet, David
and Pajdla, Tomas
and Schiele, Bernt
and Tuytelaars, Tinne",
title="Microsoft COCO: Common Objects in Context",
booktitle="Computer Vision -- ECCV 2014",
year="2014",
publisher="Springer International Publishing",
address="Cham",
pages="740--755",
isbn="978-3-319-10602-1"
}

@article{map50,
author = {Everingham, Mark and Van Gool, Luc and Williams, Christopher and Winn, John and Zisserman, Andrew},
year = {2010},
month = {06},
pages = {303-338},
title = {The Pascal Visual Object Classes (VOC) challenge},
volume = {88},
journal = {International Journal of Computer Vision},
doi = {10.1007/s11263-009-0275-4}
}

@article{hota,
author = {Luiten, Jonathon and Osep, Aljosa and Dendorfer, Patrick and Torr, Philip and Geiger, Andreas and Leal-Taixé, Laura and Leibe, Bastian},
year = {2021},
month = {02},
pages = {1-31},
title = {HOTA: A Higher Order Metric for Evaluating Multi-object Tracking},
volume = {129},
journal = {International Journal of Computer Vision},
doi = {10.1007/s11263-020-01375-2}
}

@article{mota,
author = {Bernardin, Keni and Stiefelhagen, Rainer},
year = {2008},
month = {01},
pages = {},
title = {Evaluating multiple object tracking performance: The CLEAR MOT metrics},
volume = {2008},
journal = {EURASIP Journal on Image and Video Processing},
doi = {10.1155/2008/246309}
}

@inproceedings{Ristani2016PerformanceMA,
  title={Performance Measures and a Data Set for Multi-target, Multi-camera Tracking},
  author={Ergys Ristani and Francesco Solera and Roger S. Zou and Rita Cucchiara and Carlo Tomasi},
  booktitle={ECCV Workshops},
  year={2016},
}

@inproceedings{kingma2015adam,
  title     = {Adam: A Method for Stochastic Optimization},
  author    = {Kingma, Diederik P. and Ba, Jimmy},
  booktitle = {International Conference on Learning Representations (ICLR)},
  year      = {2015}
}

@inproceedings{optuna_2019,
	title={Optuna: A Next-generation Hyperparameter Optimization Framework},
	author={Akiba, Takuya and Sano, Shotaro and Yanase, Toshihiko and Ohta, Takeru and Koyama, Masanori},
	booktitle={Proceedings of the 25th {ACM} {SIGKDD} International Conference on Knowledge Discovery and Data Mining},
	year={2019}
}

@inproceedings{convgru,
  title={Delving Deeper into Convolutional Networks for Learning Video Representations},
  author={Ballas, Nicolas and Yao, Li and Pal, Christopher J. and Courville, Aaron},
  booktitle={International Conference on Learning Representations (ICLR)},
  year={2016}
}

@inproceedings{zhang2022bytetrack,
  title={Bytetrack: Multi-object tracking by associating every detection box},
  author={Zhang, Yifu and Sun, Peize and Jiang, Yi and Yu, Dongdong and Weng, Fucheng and Yuan, Zehuan and Luo, Ping and Liu, Wenyu and Wang, Xinggang},
  booktitle={European conference on computer vision (ECCV)},
  pages={1--21},
  year={2022},
  organization={Springer}
}

@article{laumer2016compressed,
author = {Laumer, Marcus and Amon, Peter and Hutter, Andreas and Kaup, André},
year = {2016},
month = {01},
pages = {},
title = {Moving object detection in the H.264/AVC compressed domain},
volume = {5},
journal = {APSIPA Transactions on Signal and Information Processing},
doi = {10.1017/ATSIP.2016.18}
}

@inproceedings{laumer2013compressed,
  author    = {Laumer, M. and Amon, P. and Hutter, A. and Kaup, A.},
  title     = {Compressed Domain Moving Object Detection Based on H.264/AVC Macroblock Types},
  booktitle = {Proceedings of the 8th International Conference on Computer Vision Theory and Applications (VISAPP 2013)},
  pages     = {219--228},
  year      = {2013},
  address   = {Barcelona, Spain},
  publisher = {SCITEPRESS}
}

@article{Patel2014H264MotionSegmentation,
  author  = {Khushbu Patel},
  title   = {Motion Detection and Segmentation in H.264 Compressed Domain for Video Surveillance Application},
  journal = {International Journal of Engineering Research \& Technology (IJERT)},
  volume  = {3},
  number  = {4},
  year    = {2014},
  pages   = {IJERTV3IS041541},
  doi     = {10.17577/IJERTV3IS041541},
  url     = {https://www.ijert.org/research/motion-detection-and-segmentation-in-h.264-compressed-domain-for-video-surveillance-application-IJERTV3IS041541.pdf}
}

@article{Mezaris2004RealtimeCS,
  title={Real-time compressed-domain spatiotemporal segmentation and ontologies for video indexing and retrieval},
  author={Vasileios Mezaris and Yiannis Kompatsiaris and Nikolaos V. Boulgouris and Michael G. Strintzis},
  journal={IEEE Transactions on Circuits and Systems for Video Technology},
  year={2004},
  volume={14},
  pages={606-621},
  url={https://api.semanticscholar.org/CorpusID:12563363}
}

@inproceedings{bombardelli2018efficient,
  title={Efficient object tracking in compressed video streams with graph cuts},
  author={Bombardelli, Fernando and G{\"u}l, Serhan and Becker, Daniel and Schmidt, Matthias and Hellge, Cornelius},
  booktitle={2018 IEEE 20th International Workshop on Multimedia Signal Processing (MMSP)},
  pages={1--6},
  year={2018},
  organization={IEEE}
}

@article{duche2026see,
  title={See Without Decoding: Motion-Vector-Based Tracking in Compressed Video},
  author={Duch{\'e}, Axel and Chatelain, Cl{\'e}ment and Gasso, Gilles},
  journal={arXiv preprint arXiv:2602.00153},
  year={2026}
}

@ARTICLE{elkhoury,
    
AUTHOR={El Khoury, Karim  and Samelson, Jonathan  and Macq, Benoît },
           
TITLE={Deep Learning-Based Object Tracking via Compressed Domain Residual Frames},
          
JOURNAL={Frontiers in Signal Processing},
          
VOLUME={Volume 1 - 2021},
  
YEAR={2021},
  
URL={https://www.frontiersin.org/journals/signal-processing/articles/10.3389/frsip.2021.765006},
  
DOI={10.3389/frsip.2021.765006},
  
ISSN={2673-8198},
}
\end{document}